\documentclass[11pt,a4paper]{article}
\usepackage[hyperref]{emnlp2020}
\usepackage{times}
\usepackage{latexsym}
\usepackage{url}
\usepackage{soul}
\usepackage{graphicx}
\usepackage{booktabs}
\usepackage{siunitx}
\usepackage{amsmath}
\usepackage{amssymb}
\usepackage{bm}
\usepackage{xspace}
\usepackage{eucal}
\usepackage{calligra}
\usepackage{multicol}
\usepackage{lipsum}
\usepackage{algorithm}
\usepackage{algorithmic}
\usepackage{color}
\usepackage{enumitem}
\setlist{nosep}

\newcommand{\RR}{\mathbb R}

\usepackage{microtype}

\aclfinalcopy %

\title{Multi-Pass Transformer for Machine Translation}

\author{Peng Gao$^{1,2}$\thanks{*This work was done while Peng Gao and Shijie Geng were interns at MERL} \quad Chiori Hori$ ^1$ \quad Shijie Geng$ ^{1,3*}$ \quad Takaaki Hori$ ^1$ \quad Jonathan Le Roux$ ^1$\\
  $^1$Mitsubishi Electric Research Laboratories (MERL) \\
  $^2$Chinese University of Hong Kong \\
  $^3$Rutgers University\\
  {\tt\normalsize \{chori, thori, leroux\}@merl.com}}
  
\date{}

\allowdisplaybreaks

\begin{document}
\maketitle
\begin{abstract}
In contrast with previous approaches where information flows only towards deeper layers of a stack, we consider a multi-pass transformer (MPT) architecture in which earlier layers are allowed to process information in light of the output of later layers. To maintain a directed acyclic graph structure, the encoder stack of a transformer is repeated along a new multi-pass dimension, keeping the parameters tied, and information is allowed to proceed unidirectionally both towards deeper layers within an encoder stack and towards any layer of subsequent stacks.
We consider both soft (i.e., continuous) and hard (i.e., discrete) connections between parallel encoder stacks, relying on a neural architecture search to find the best connection pattern in the hard case. We perform an extensive ablation study of the proposed MPT architecture
and compare it with other state-of-the-art
transformer architectures. Surprisingly, Base Transformer equipped with MPT can surpass the performance of Large Transformer on the challenging machine translation En-De and En-Fr datasets. In the hard connection case, the optimal connection pattern found for En-De also leads to improved performance for En-Fr. %
\end{abstract}

\section{Introduction}
In recent years, we have witnessed the evolution of neural architectures from recurrent neural network (RNN) to long short-term memory (LSTM)~\cite{hochreiter1997long}, convolutional neural network (CNN), and transformer~\cite{vaswani2017attention}. Thanks to the increasing ability to capture contextual information due to better architectures, performance on language understanding and generation tasks has dramatically improved. Convolutional sequence model~\cite{gehring2017convolutional} and Transformer have been the most popular architectures for language representation learning.

\begin{figure}[t!]
\begin{center}
\includegraphics[width=0.8\linewidth,trim={0cm 0cm 0cm 0cm}]{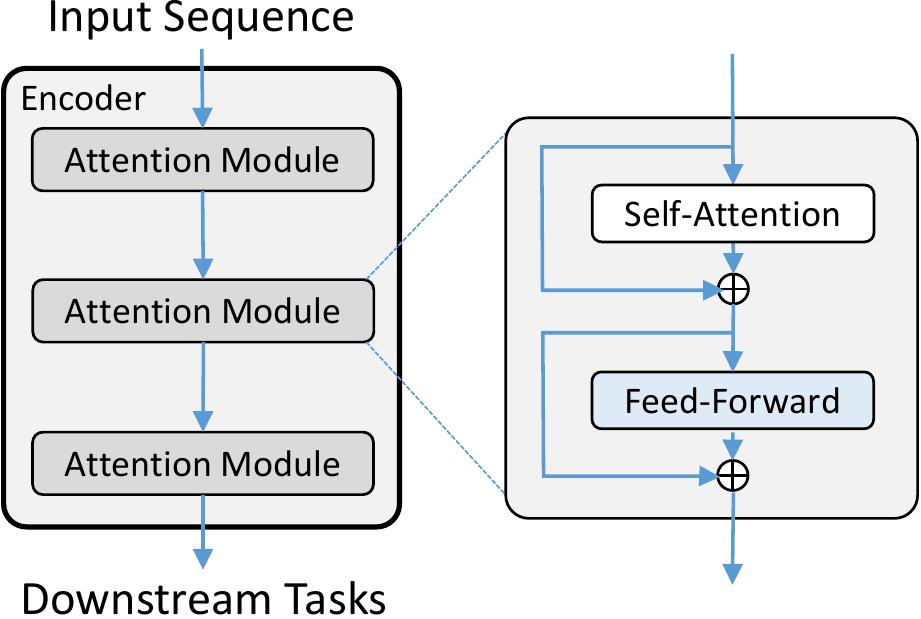}
\end{center}
\caption{Sequential Transformer architectures have been widely adopted as encoders for many \emph{downstream} tasks. Each attention module contains one self-attention and one feed-forward network module.}
\label{Sequential}
\end{figure}

However, transformer and convolutional sequence model simply employ a sequential architecture in which information flows unidirectionally from earlier layers to later layers within the architecture, 
and downstream tasks only have indirect access to early features through multiple layers
as shown in Fig.~\ref{Sequential}. We hypothesize that
earlier layers may benefit from processing information in light of the output of later layers, and that later layers may also benefit from more direct connections with earlier layers, as various layers may contain complementary information useful to down-stream tasks such as language understanding and decoding.
Introducing connections from later layers back to earlier layers is however difficult, as the directed acyclic graph structure of the models needs to be maintained in order to effectively train them and use them to perform inference. 
Regarding the introduction of more direct connections from earlier layers to later layers, a naive approach %
is to use dense connections~\cite{huang2017densely} for all modules, but this results in a significant increase in model size and memory usage, making it difficult to deploy the learned model. Besides, since the dense connection increases the number of parameters, care must be taken to fairly compare performance with other models.

To solve these problems, we consider a multi-pass transformer (MPT) architecture in which the encoder stack of a transformer is repeated along a new multi-pass dimension, keeping the parameters tied, and information is allowed to proceed unidirectionally both towards deeper layers within an encoder stack and towards any layer of subsequent stacks. Subsequent stacks can thus have a chance to ``re-process'' an input in light of features obtained from later layers in a previous stack, and layers closer to the output of the stack may be more directly connected to earlier layers in the previous stack. Weights are shared between the same layers in different stacks, so that the number of parameters is not increased.

\begin{figure}[t!]
\begin{center}
\includegraphics[width=0.99\linewidth,trim={0cm 0cm 0cm 0cm}]{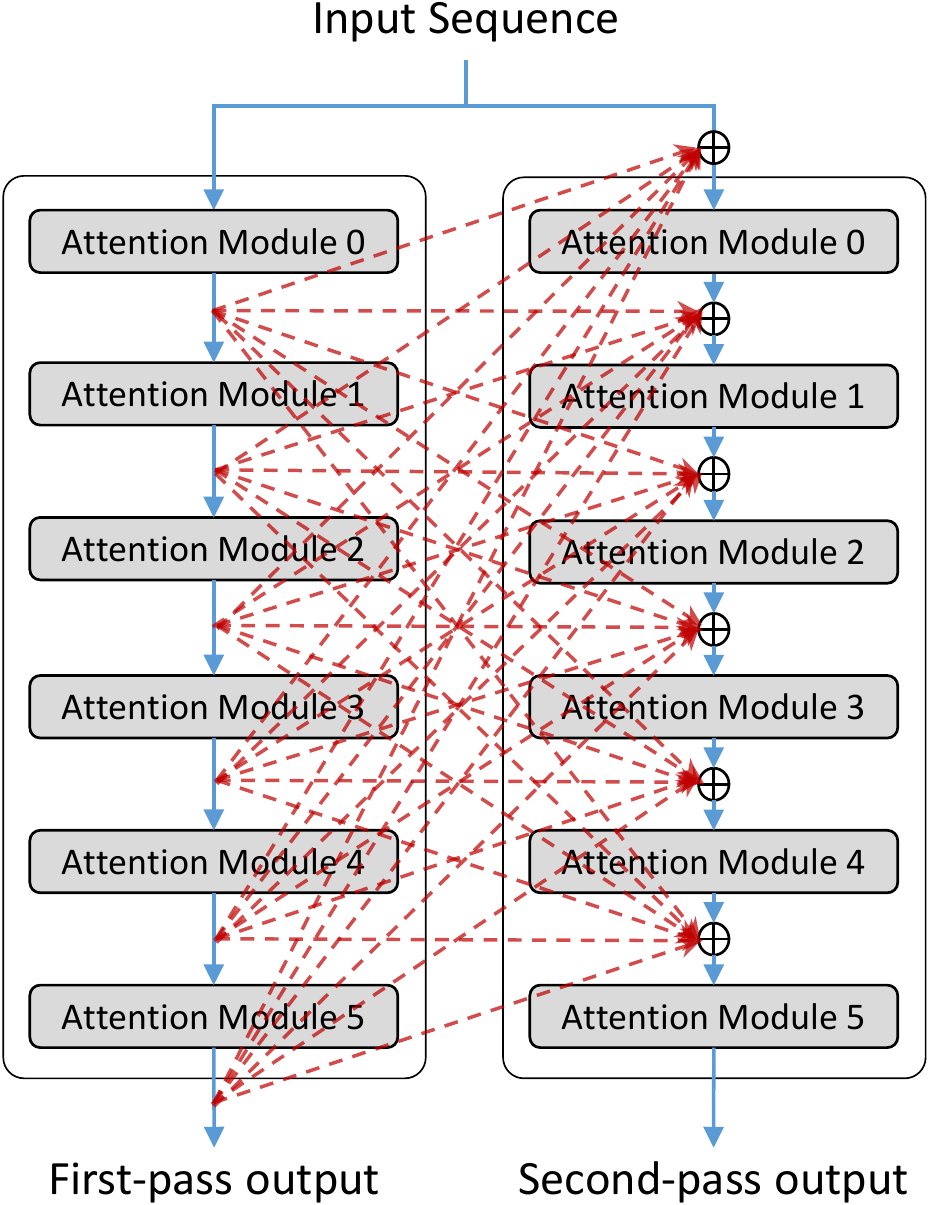}
\end{center}
\caption{Illustration of our proposed multi-pass transformer. The network consisting of the modules on the left side, referred to as the first-pass encoder stack or the inner stack, is copied to create a second-pass encoder stack, also referred to as the outer stack, sharing the same weights with the inner stack. The output of each layer in the inner stack is fed as input into one or more of the layers in the outer stack by adding it to their other inputs, similarly to a residual or skip connection. All possible such connections are here shown. In the soft connection case, a weight is learned for the residual connection between each layer pair. In the hard connection case, each layer can only be involved in a single residual connection, and the optimal connection pattern between inner and outer networks is found by random search. Information is propagated from top to bottom then from left to right.}
\label{Flow}
\end{figure}

The MPT architecture is illustrated in Fig.~\ref{Flow} for the case of a two-pass transformer. A single encoder stack, which can be considered as the first-pass encoder, is cloned to a second-pass encoder with identical weights, and connections between all layers of the two stacks are considered based on residual-like connections, i.e., the output of a layer in the first stack can be added to the input of a layer in the second stack. For simplicity, we will also refer to the first stack and its layers as the ``inner stack'' and ``inner layers'', and the second stack and its layers as the ``outer stack'' and ``outer layers''.
We consider two particular cases for these connections: in the soft connection case, a weighted combination of the outputs of all inner layers is added to the input of each outer layer, where the weights are learned during training together with the network parameters; in the hard connection case, the output of each inner layer is added to the input of a single outer layer, which can only receive information from that inner layer. The soft connection case adds a negligible amount of parameters to the network, while the hard connection case requires performing a discrete architecture search. We use random neural architecture search to find the best connection pattern, motivated by previous work exploring such methods to obtain a good CNN structure for image recognition~\cite{xie2019exploring} .

In this paper, we only consider modifications of the encoder part of the network, and leave the exploration of similar modifications to the decoder part to future work.

Our contribution can be summarized as follows:
\begin{itemize}[leftmargin=*]%
  \item We propose a novel multi-pass transformer which allows a transformer encoder to process an input in light of features obtained at various, including later, stages of the processing. MPT can outperform Large Transformer by reusing features from Base Transformer without increasing the number of parameters.
  \item By training for downstream tasks using either the output of the first or the second pass encoder, we simultaneously train two network architectures with different computation/performance trade-offs.
  \item We perform an extensive ablation study to validate the design of our proposed multi-pass transformer.
  \item In contrast with other architecture search methods which use recurrent controller~\cite{zoph2016neural}, complex reinforcement learning and evolution algorithm~\cite{so2019evolved}, we design a simple random architecture search on a carefully designed search space. We can outperform the Evolved Transformer that was searched by Google in a much larger search space, thus validating the effectiveness of our designed search space. 
  
\end{itemize}

\section{Related Work}
\subsection{Neural Architecture}
Convolution neural network, recurrent neural network, long short-term memory, and transformer have been the fundamental building blocks for modern neural architectures. RNN utilizes a recurrent memory for temporal information propagation. CNN applies a learnable sliding window for local neighborhood information aggregation. Transformer utilizes fully-connected self-attention to learn pairwise information relationships. In recent years, self attention has been applied to representation learning~\cite{devlin2018bert}, vision language reasoning~\cite{Gao_2019_CVPR, Gao_2019_ICCV}, vision relationship learning~\cite{hu2018relation}, and video temporal information aggregation and reasoning~\cite{wang2018non,geng2020character,geng2020spatio}. Dense networks and residual connections~\cite{he2016deep} have been popular multi-path connection patterns for preventing the vanishing or exploding gradient problem. %
Besides residual and dense connections, Xception~\cite{chollet2017xception}, dual path~\cite{chen2017dual}, and FPN~\cite{lin2017feature} have also been quite popular. Most architectures in computer vision adopt a multi-path approach for efficient information flow. However, sequential architectures with limited information flow are typically adopted in natural language processing (NLP). The motivation for our proposed multi-pass transformer is to bring insights from the multi-path architecture design popular in computer vision to the transformer architecture originally introduced in NLP, and to go beyond the typically unidirectional multi-path architectures through a multi-pass extension.

\subsection{Neural Architecture Search}
AutoML~\cite{zoph2016neural} firstly proposed neural architecture search (NAS) to find optimal structures for image and language models. An RNN controller with a reward function is utilized to generate an optimal structure. After training 12,800 architectures on CIFAR,  state-of-the-art performance is achieved. The evolved transformer~\cite{so2019evolved} applied evolution search on a specifically designed space with a combination of self-attention and convolutional neural networks for machine translation. After searching, the evolved transformer showed that a model with a dual path consisting of convolutional neural network and self-attention can achieve better performance than a pure self-attention model. Reinforcement learning and evolutionary approaches have been widely used in many neural architecture search papers, and are believed to be better than random search. However, the randomly wired network~\cite{lin2017feature} showed that a random search on a well-designed space can outperform state-of-the-art models found by evolutionary algorithm.

\subsection{Multi-Stage Fusion in Computer Vision and Natural Language Processing}
Combination of residual and dense connections is the most popular multi-stage fusion mechanism applied in computer vision and natural language processing. Multi-stage fusion uses skip connections and multi-stage feature fusion to perform efficient information fusion of lower and deeper features. This can somewhat alleviate the vanishing or exploding gradient problem. Multi-stage fusion has been widely used as a feature pyramid in object detection. Recently, multi-stage feature fusion has also been applied to machine translation. Simple fusion mechanisms such as concatenation, addition, and recurrent fusion have been applied to fuse multi-stage features~\cite{wang2019exploiting}. However, the model cannot capture multi-stage information due to the limited capacity of such simple fusion mechanisms. Dynamic layer aggregation~\cite{dou2019dynamic} recently proposed a novel routing-by-agreement algorithm which can aggregate information from multiple layers by an expectation-maximization algorithm, achieving the state-of-the-art on machine translation.

\section{Proposed Method}
\subsection{Transformer for Machine Translation}
Transformer architectures have recently been widely used for sequence generation tasks. 
The typical structure consists of encoder and decoder networks, which have deep feed-forward architectures including repeated blocks of self-attention and feed-forward layers with residual connections~\cite{he2016deep} and layer normalization~\cite{ba2016layer}.
The decoder network also features a source attention layer in each block to read the encoder's output. In this paper, we focus on modifying the encoder structure only, leaving the modification of the decoder to future work. Given a language pair, our aim is to translate a source sentence to a target sentence. The source sentence is tokenized by byte-pair encoding (BPE) and then transformed by a word embedding layer into a representation consisting of $L$ vectors of dimension $C$, where $L$ is the sentence length and $C$ the word embedding dimension. The position of each word is encoded into a position embedding space and added to the word embedding. The target sentence is similarly encoded into a word embedding representation, shifted to the right in order to force the decoder to only look at past outputs to predict the next output during training.

The continuous representation $S\in \RR^{L\times C}$ of the source at the input of a self-attention module is translated via linear transforms into query, key, and value vector sequences, denoted respectively as $S^Q$, $S^K$, and $S^V$. By using an attention value between key and query, each word in the source can aggregate information from other words using self-attention. The attention outputs can be calculated as:
\begin{equation}
    \mathrm{Attention}(Q,K,V) = \mathrm{softmax}\Big(\frac{QK^T}{\sqrt{d_K}}\Big)V,
\end{equation}
where $Q$, $K$, and $V$ denote the query, key, and value vector sequences, and the attention value is modulated by the square root of the feature dimension $d_K$.
After the self-attention module aggregates information from other words, a position-wise feed-forward network layer (FFN) is used to fuse that information. Each $k$-th attention module of the transformer encoder is a combination of one self-attention layer and one FFN layer, processing the input $S^\text{in}_k$ as follows:
\begin{align}
    S^\text{mid}_k &= \mathrm{Attention}(S^{\text{in},Q}_k, S^{\text{in},K}_k, S^{\text{in},V}_k), \\
    S^\text{out}_k &= \mathrm{FFN}(S^\text{mid}_k).
\end{align}
where $S^\text{mid}_k$ stands for the intermediate feature inside the $k$-th module, and $S^\text{out}_k$ for the final output after FFN. These two notions of output features for a layer, intermediate and final, will be considered in the introduction of our proposed multi-pass transformer.
Several modules are stacked to acquire a good representation of the source sentence, and self-attention is typically implemented using multi-head attention. 

The decoding stage is similar to the encoding stage, except that self-attention is performed on each target sentence's embedding representation $T$, and followed by source-attention and FFN. One layer of the decoding stage performs the following processing, where SA stands for self-attention, and $S_{N-1}$ denotes the output of the last encoder layer:
\begin{align}
    T^{\text{SA},Q} &= \mathrm{Attention}(T^Q, {T}^K, {T}^V),\\
    T^{\text{out}} &= \mathrm{FFN}(\mathrm{Attention}(T^{\text{SA},Q}, {S}^{K}_{N-1}, {S}^{V}_{N-1})).
\end{align}

We use the same word embedding transformation for the encoder and decoder. After getting the representation for the next word in the decoder, we use cross-entropy loss between the decoder word embedding and the learned representation to optimize the transformer model. We denote as $\mathcal{L}(S_{N-1},T)$ the loss computed from the output $S_{N-1}$ of the encoder, with target sequence $T$.

\subsection{Multi-Pass Transformer}
The core idea of the multi-pass transformer is feature reuse, namely feeding the output of inner layers from the first-pass encoder into the input of outer layers from a second-pass encoder. By feeding, we here mean that the output of a given inner layer is added to the original input of an outer layer without any transformation (other than potentially weighting), akin to a form of residual connection.
The first-pass and second-pass networks share parameters, and the expanded architecture thus has the same number of parameters as the base transformer architecture. Arguably the simplest connection pattern within such an architecture is that in which an output of the $k$-th attention module in the first-pass encoder is added to the input of the $k$-th attention module in the second-pass encoder. We use this setup as our default MPT architecture. 
We consider in this paper two types of connections, soft and hard, which both include this simplest pattern as a particular case.

\textbf{Soft Connections:} In the soft connection case, a weighted combination of the outputs of all inner layers is added to the input of each outer layer, where the weights are learned during training together with the network parameters. 
The output of the $k$-th outer layer, $\tilde{S}_k^\textrm{out}$, can be computed as
\begin{align}
\tilde{S}_k^\textrm{out} = \textrm{AttModule}\Big(\tilde{S}_{k-1}^\textrm{out} + \sum_{j=0}^{N-1} \alpha_{kj} S_j^\textrm{out}\Big),
\end{align}
where $\textrm{AttModule}(\cdot)$ denotes the attention module consisting of self-attention and feed-forward networks, $S_j^\textrm{out}$ is the output of the $j$-th inner layer,
and $\alpha_{kj}$ represents a weight for the connection from the $j$-th inner layer to the $k$-th outer layer.
We compute the connection weight via softmax as $\alpha_{kj}=\exp(w_{kj})/\sum_{j}\exp(w_{kj})$ with learnable parameters $w_{kj}$, to enforce $0 \leq \alpha_{kj} \leq 1$ and $\sum_j \alpha_{kj}=1$. 
Our default architecture, with the simplest pattern mentioned above, corresponds to weights $\alpha_{kj}=\delta_{kj}$, where $\delta$ denotes the Kronecker delta.
Note that this approach adds $N^2$ parameters (with $N=6$ in our experiments) to the whole model, but this is negligible compared to the typical model size.

\begin{figure}[t!]
\begin{center}
\includegraphics[width=0.90\linewidth,trim={0cm 0cm 0cm 0cm}]{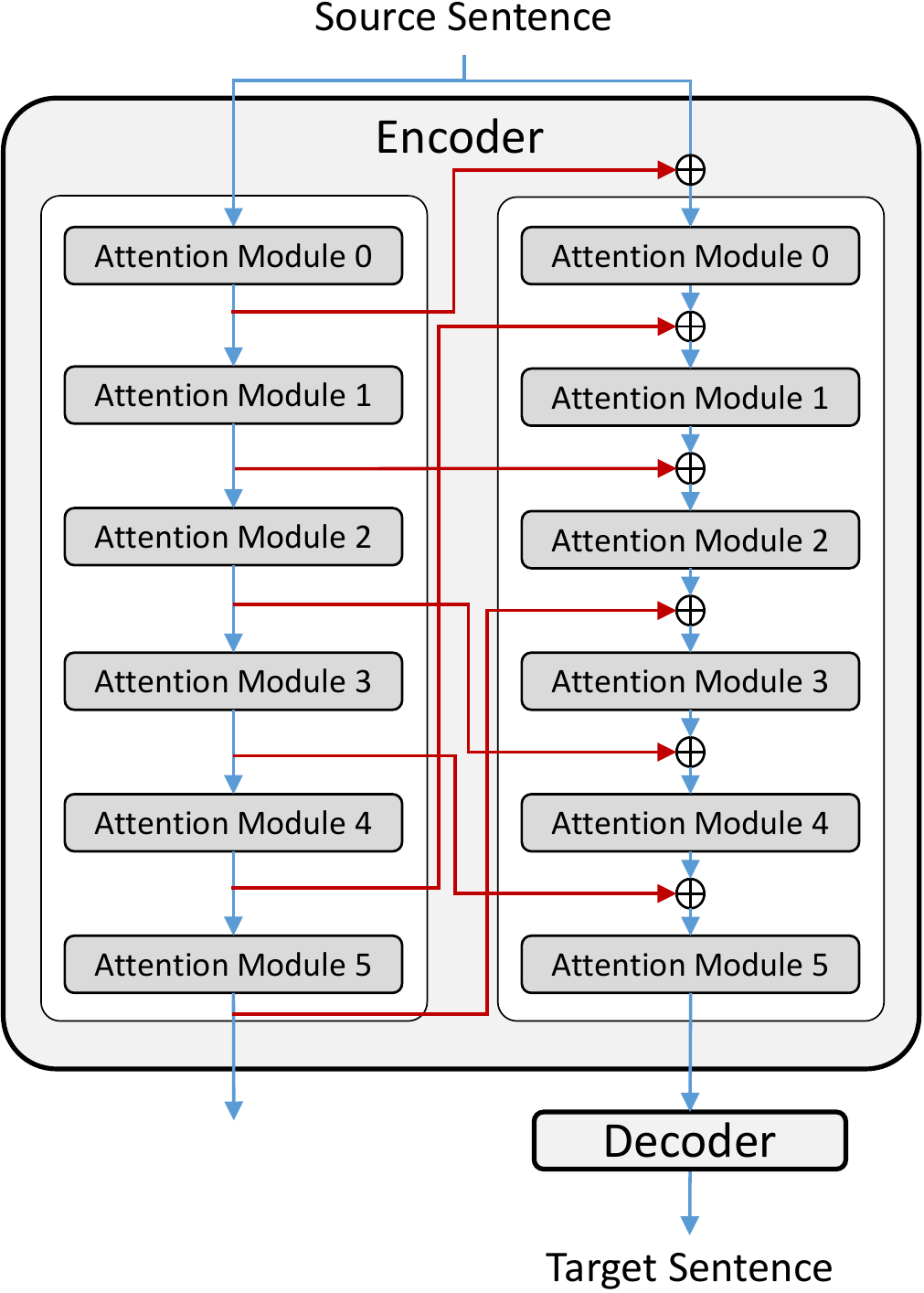}
\end{center}
\caption{Best searched model for the hard connection case with $N=6$ attention layers, $[0, 4, 1, 5, 2, 3]$.}
\label{best}
\end{figure}

\textbf{Hard Connections:} In the hard connection case, we consider configurations in which the output of one layer in the first-pass encoder is added to the input of one layer in the second-pass encoder, with the constraint that no two outputs can be added to the same input. This reduces the search space to the set of permutations on $\{0,\dots,N-1\}$, where $N$ denotes the number of layers in the inner network, resulting in a search space of size $N!$, down from $N^N$ without the above constraint. We denote the $i$-th hard MPT architecture using the image $[\tau^{(i)}_0,\dots,\tau^{(i)}_N]$ of the sequence $[0,\dots,N-1]$ via the associated $i$-th permutation. In the $i$-th hard MPT architecture, the output of layer $\tau^{(i)}_k$ in the first-pass encoder is added to the input of layer $k$ in the second-pass encoder. For example, for a base transformer architecture with $N=6$ layers, the MPT model denoted as $[0,4,1,5,2,3]$  is shown in Fig.~\ref{best}. Input to outer layer $0$ comes from inner layer $0$, that of outer layer $1$ comes from inner layer $4$, that of outer layer $2$ comes from inner layer $1$, and so on. Our default architecture, with the simplest pattern mentioned above, is another example of hard MPT architecture, which is denoted as $[0,1,2,3,4,5]$. 
Because hand-designed connection patterns are likely not to be optimal because of the lack of theory to predict the performance of a model based on its architecture, we adopt architecture search to find an optimal architecture in the hard MPT case, as explained in more details in Section~\ref{sec:search}. 

\textbf{Training procedure:} Both for the soft connection architecture and each hard connection architecture, we train the model by minimizing, for an input sequence $S$ and a target sequence $T$, the objective function $\mathcal{L}(\tilde{S}_{N-1},T)$ obtained by applying the decoder to the output $\tilde{S}_{N-1}$ of the second-pass encoder. Note that we also explored optimizing the network such that the outputs of both the first-pass and second-pass encoders could be used for downstream tasks, by minimizing the sum of $\mathcal{L}(S_{N-1},T)$ and $\mathcal{L}(\tilde{S}_{N-1},T)$ or by randomly minimizing either. Performance of the second-pass encoder output remained roughly the same in all cases. This could be useful in applications where a system may need to switch between regimes with low and high computation costs (keeping the same number of parameters), in which case it could adaptively use the output of the first-pass or second-pass encoders.

\subsection{Routing patterns between layers}
\label{sec:routing_patterns}

\begin{figure}[t]
\begin{center}
\includegraphics[width=.99\linewidth,trim={0.0cm 0cm 0.1cm 0cm}]{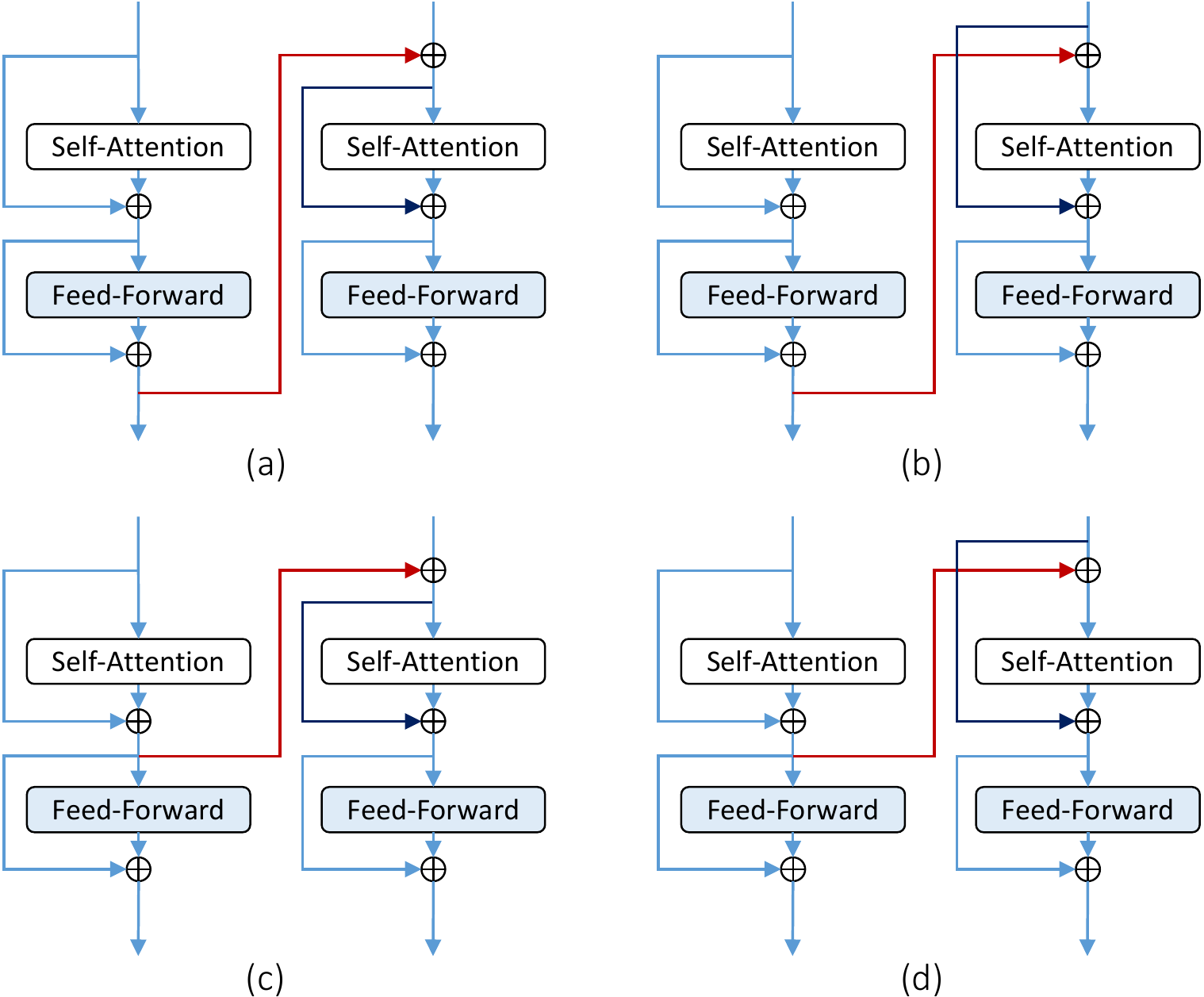}
\end{center}
\caption{Four different routing patters (a), (b), (c) and (d) depending on whether a feature from an inner layer is taken at the output or from the middle of a transformer module, and whether it is added to the input of an outer layer before or after the residual connection.}
\label{connection}
\end{figure}

As can be seen in Fig.~\ref{connection}, we consider four different routing patterns to infuse features from an inner layer to an outer layer. In pattern (a), the output feature of a transformer module is added into an outer network layer before the residual connection. By connecting before the residual, the information propagates to all the following layers. In pattern (b), the feature from the inner layer is infused into the outer layer after the residual connection. By doing so, information cannot propagate as directly to the following layers. In pattern (c), instead of adding the output of the feed-forward neural network into the outer network layer, we directly feed the information after self-attention into the outer network layer before the residual connection. In pattern (d), the information after self-attention is added into the outer network layer after the residual connection. The performance of (c) and (d) is expected to be worse than that of (a) and (b), because the information after self-attention has not been fully fused yet, and the model's learning ability is thus likely to be decreased. We compare these routing patterns in Section~\ref{sec:ablation}

\subsection{Architecture Search for Hard MPT}
\label{sec:search}
For the base transformer architecture with 6 layers shown on the left side in Fig.~\ref{best}, there are 720 associated hard MPT architectures respecting the constraint we imposed, versus 46,656 without imposing the constraint. In previous research, reinforcement learning and evolution have been applied for architecture search. Here,  we use random search for its simplicity and efficiency. A naive approach would be a random search over all architectures. However, there is only one connection swap between neighboring structures in the search space: we hypothesize that one swap will in general not significantly affect the performance. By using this prior knowledge, we perform a coarse random search, and forgo search on neighbours of structures with bad results. We instead perform fine-grained search around models with good performance.

\section{Experiment}
\subsection{Experiment Setup}
We follow the same experiment setup previously used for the Transformer~\cite{vaswani2017attention}. 
We perform an ablation study on the WMT 2014 English-German (En-De) dataset, which consists of 4.5 million sentence pairs. Byte-pair encoding was applied to generate a dictionary containing 32,000 tokens. We also evaluate our models on the significantly larger WMT 2014 English-French dataset, consisting of 36 million sentence pairs.
Sentences with approximately the same length are sampled into the same batch. A single Titan RTX GPU with 24 GB of memory is used to train each base transformer model during model search architecture. By utilizing single GPU training during architecture search, we can increase the overall training speed without communication between different GPUs. Training one base transformer requires about 48 hours. With 32 GPUs running in parallel, we were able to train 224 models in two weeks. All models shared the same set of hyper-parameters. Learning rate was set to 0.0008, with 10,752 tokens in each batch with a gradient accumulation step of 8 to simulate 8-GPU parallel training. Besides, weight decay was 0 with Adam optimization whose hyperparameters were set as (0.9, 0.98). We utilized the same learning schedule as other papers by using the inverse square root. Dropout ratio was set to $0.1$ for all experiments with label smoothing $0.1$. We utilized fp16 for training acceleration. Our Base transformer contained 6 layers of self-attention modules with 512 hidden dimensions by default. During evaluation, we averaged the model parameters of the last 5 checkpoints and ran a grid search for optimal beam size and length penalty in beam search. We found that beam size of 4 and length penalty of 0.2 were the optimal setup for our searched model.

\subsection{Ablation Study}
\label{sec:ablation}

We first validate various design choices in our proposed architectures. %
We compare our results with the performance of the baseline Base Transformer architecture. This model obtained a BLEU score of 27.3 in the original paper~\cite{vaswani2017attention}, while our reimplementation obtained 27.6.

\noindent \textbf{Comparing routing patterns:}
As shown in Fig.~\ref{connection}, several routing patterns can be considered for reusing features from inner layers in outer layers. 
We utilize our default MPT architecture, which corresponds to the hard MPT architecture $[0, 1, 2, 3, 4, 5]$, and test its performance on routing patterns (a), (b), (c), and (d). As we hypothesised in Section~\ref{sec:routing_patterns}, Table~\ref{tab:result_ablation} shows that infusing information before the initiation of the residual connection leads to better performance. Furthermore, feeding the information obtained after the feed-forward layer is better than the feature obtained after the self-attention layer. We hypothesize that the feed-forward layer better fuses the information aggregated from other words. For all other experiments going forward, we use routing pattern (a).

\begin{table}[t]
\centering
\resizebox{\columnwidth}{!}{
\begin{tabular}{lc}
\toprule
\textbf{Model} & \textbf{BLEU}  \\
\midrule 
Base Transformer \cite{vaswani2017attention} & 27.3 %
\\
Base Transformer (our implementation) & 27.6 %
\\
\midrule
MP Transformer [0,1,2,3,4,5] & 28.1 \\
\midrule
MP Transformer [0,1,2,3,4,5](a) & 28.1\\
MP Transformer [0,1,2,3,4,5](b) & 27.8\\
MP Transformer [0,1,2,3,4,5](c) & 27.5\\
MP Transformer [0,1,2,3,4,5](d) & 27.3\\
\midrule
MP Transformer [5,2,0,4,1,3] (Worst) & 27.6\\
MP Transformer Avg.&  27.9\\
MP Transformer [0,4,1,5,2,3] (Best) & 28.4\\
MP Transformer [Soft Connections] & 28.4 \\
\bottomrule
\end{tabular}}
\caption{Ablation study of our proposed approach on WMT 14 En-De. The proposed multi-pass transformer does not increase the number of parameters with respect to the original architecture (except in the soft connection case, but only by a negligible amount). All models thus share the same number of parameters. }
\label{tab:result_ablation}
\end{table}

\noindent \textbf{Effectiveness of Random NAS for hard connections:}
As shown in Table~\ref{tab:result_ablation}, random search can generate hard MPT architectures with significant variance, which validates the effectiveness of our carefully designed search space. The performance difference between the best and worst model is 0.8 (the best model obtains 28.4 while the worst model obtains 27.6), which is large. The best model is the hard MPT architecture with connection pattern $[0, 4, 1, 5, 2, 3]$, which is illustrated in Fig.~\ref{best}. By analysing good and bad models, we notice the following factors may influence the performance. As seen from the searched network, features from deeper layer in the inner network should be added to shallow features in the outer network, which can increase the longest information path existing in the model. However, we notice that infusing the final layer of the inner network to the first layer of the outer network results in significantly deteriorated performance; we hypothesize that the existence of too long a path results in training issues due to gradient instability. Conversely, performance also seems to improve when features from shallow layers in the first network are directly linked to deeper layers in the outer network. By doing so, our multi-pass transformer has both long path and short path information flows, which may be the reason for its good performance.

\begin{figure}[t!]
\begin{center}
\includegraphics[width=0.99\linewidth,trim={0cm 0cm 0cm 0cm}]{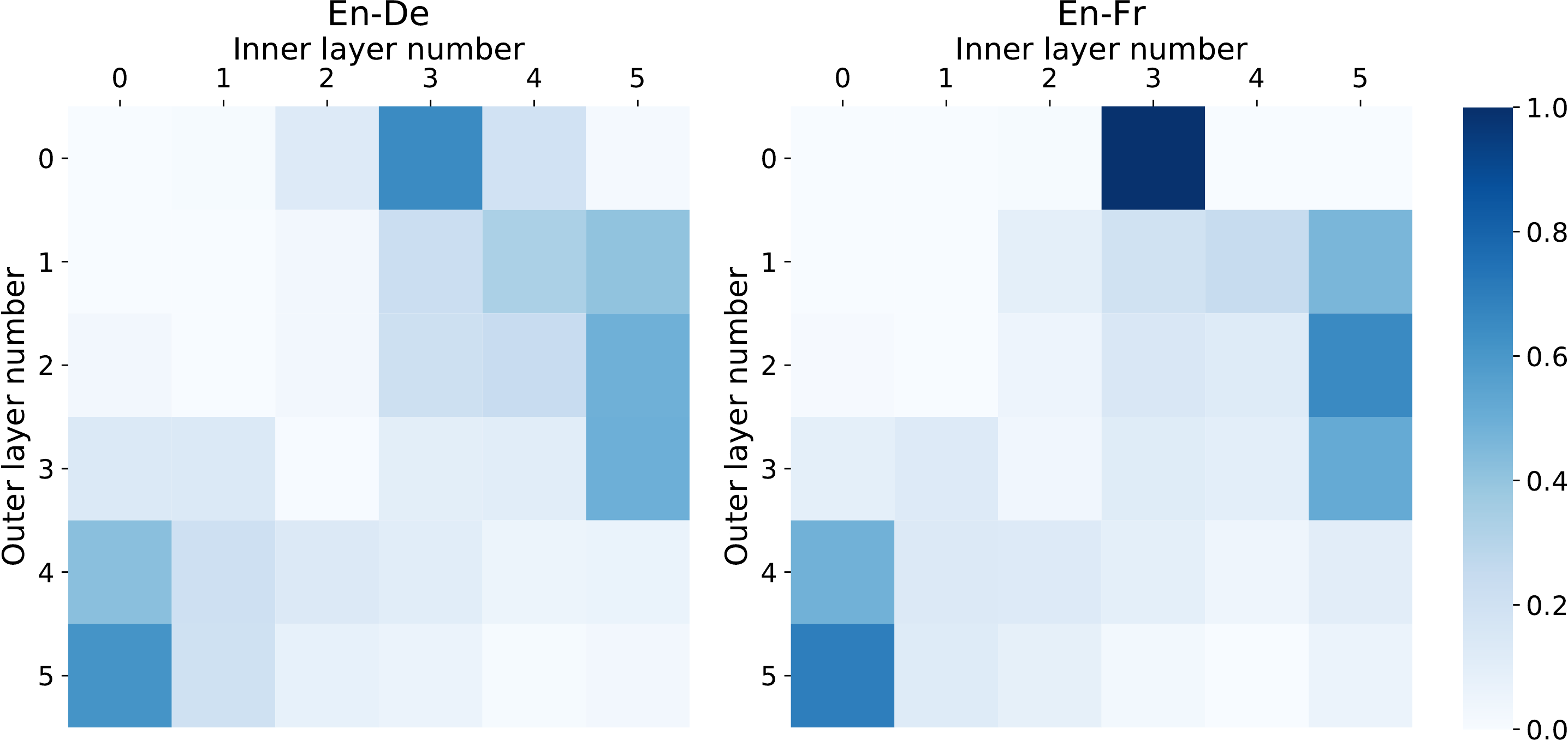}
\end{center} %
\caption{Soft connection weights obtained on the En-De and En-Fr datasets.} %
\label{fig:soft_weights}
\end{figure}

\noindent \textbf{Soft connections:} Interestingly, we can see that the performance of the MPT architecture with soft connections exactly matches that of the best hard MPT architecture on the En-De dataset, with 28.4. The optimal weights obtained on the En-De and En-Fr datasets are shown in Fig.~\ref{fig:soft_weights}. We note that these optimal weights fit very well with the above analysis regarding the performance trends in the hard connection case.

\begin{table*}
\centering
\setlength{\tabcolsep}{5.5pt}
\sisetup{table-format=2.1,round-mode=places,round-precision=1,table-number-alignment = center,detect-weight=true,detect-inline-weight=math}
\scalebox{0.9}{
\begin{tabular}{rSSSS}
\toprule
\multicolumn{1}{c}{\textbf{Model}} & \multicolumn{2}{c}{\textbf{BLEU}}  & \multicolumn{2}{c}{\textbf{Params $\times 10^6$}} \\
\cline{2-5}
& \text{EN-DE} & \text{EN-FR} & \text{EN-DE} & \text{EN-FR} \\
\midrule 
Base Transformer (BT) \cite{vaswani2017attention} & 27.30 & 38.1 & 61.2 & 111.4 \\
Large Transformer \cite{vaswani2017attention} & 28.40 & %
41.8 & 213.0 & 310.9 \\
\midrule
The Evolved Transformer (ICML 2019) \cite{so2019evolved} & 28.20 & 41.3 & 64.1 & 221.2 \\
\midrule
Sentential Context Max Pooling (ACL 2019) \cite{wang2019exploiting} & 27.58 & & 106.9 & \\
Sentential Context Attention \cite{wang2019exploiting} & 27.81 & & 107.9  \\
Deep Sentential Context RNN \cite{wang2019exploiting} & 28.33 & 40.27 & 114.3 & 114.3 \\
\midrule
Linear Combination + BT (ACL 2019) \cite{dou2019dynamic} & 27.73 & & 102.7 & \\
Dynamic Combination + BT \cite{dou2019dynamic} & 28.33 & & 113.2 & \\
Dynamic Routing + BT ~\cite{dou2019dynamic} & 28.22 & & 125.8 & \\
EM Routing + BT ~\cite{dou2019dynamic} & 28.81 & & 144.8 & \\
\midrule
MSC Base Transformer (BT)~\cite{wei2020multiscale} & 27.68 & & 73.0 & \\ 
\midrule
6-layer Base Transformer (BT) & 27.6 & & 61.2 & \\ 
6+6-layer (tied weihts) Base Transformer (BT) & 27.8 & & 61.2 & \\ 
12-layer Base Transformer (BT) & 28.5 & & 79.8 & \\ 
MP Transformer [0,4,1,5,2,3] (Proposed) & 28.40 & 41.80 & 61.2 & 111.4 \\
MP Transformer [Soft Connections] (Proposed) & 28.43 & 41.56 & 61.2 & 111.4 \\ 
\bottomrule
\end{tabular}
}
\caption{Comparison with other state-of-the-art methods on English WMT14 En-De and En-Fr datasets. To compare with simple deep layer structures, a fully unconstrained
12-layer BT and with a model simply concatenating two 6-layer BTs with tied weights (but without feeding the original input to the 7th layer nor allowing extra direct connections between layers 1-6 and 7-12 as we propose) were tested for En-De.}
\label{tab:result_sota}
\end{table*}

\subsection{Comparison with State of the Art}
In this section, we compare our best soft and hard connection architectures with state-of-the-art extensions of the base transformer with similar settings on both the En-De and En-Fr datasets. For the En-Fr dataset, we train the soft MPT architecture, as well as the hard MPT architecture with the best connection pattern searched on the En-De dataset. Results are reported in Table~\ref{tab:result_sota}. Interestingly, both our architectures outperform all other models while using less parameters. We also observe that the best hard connection pattern found on En-De generalizes well to En-Fr, and even outperforms the soft connection architecture on that dataset. We now give more detailed accounts regarding how each of the other models relates to our proposed models.

\noindent\textbf{The Evolved Transformer~\cite{so2019evolved}:}
The evolved transformer performs architecture search on a much larger search space than us by using an evolutionary algorithm. An extensive architecture search is performed on the size of the self-attention heads, the number of layers, different cascades between convolution and self-attention, and dense-residual fusion and  architecture search is performed jointly on the encoder and decoder. The evolved transformer involves a much larger search space than our multi-pass transformer. In the proposed hard MPT architecture, we only perform random search on a much more restricted but carefully designed search space. The soft MPT architecture does not even need architecture search, directly estimating the best continuous connection pattern. Both our architectures can achieve better performance than the evolved transformer as shown in Table ~\ref{tab:result_sota}, which validates the effectiveness of our proposed multi-pass transformer. 

\noindent\textbf{Sentential Context Transformer  ~\cite{wang2019exploiting}:}
The sentential context transformer proposes to fuse features from all layers in the encoder network by a simple fusion approach like addition, recurrent fusion, concatenation, or attention. Due to the simplicity of the proposed approaches, only partial information in the features from all layers can be captured. Furthermore, operators like concatenation and recurrent neural network significantly increase the number of parameters as seen in Table~\ref{tab:result_sota}. Our proposed models can achieve much better performance than the sentential model with much fewer parameters as shown in Table ~\ref{tab:result_ablation}. %

\noindent\textbf{Dynamic Layer Aggregation (DLA)~\cite{dou2019dynamic}:}
DLA shares the same design concept with the sentential context transformer by utilizing multi-layer information fusion, but proposes a much more efficient mechanism for layer information aggregation by utilizing the expectation-maximization (EM) algorithm. The EM-based routing-by-agreement algorithm achieves significantly better performance. However, the model has almost double the number of parameters compared to our models as shown in Table~\ref{tab:result_sota}. We consider dynamic layer aggregation as an orthogonal direction with respect to our proposed MPT, and we plan to combine the two approaches in the future. 

\noindent\textbf{Multiscale Collaborative Deep Models (MSC)~\cite{wei2020multiscale}:}
In order to easily optimize parameters when training very deep NMT models, MSC applied a block-scale collaboration mechanism exploiting shortcut connections propagating gradients from the lower levels of the encoder to the decoder. To enhance source representations with spatial dependencies by contextual collaboration, Gated Recurrent Units (GRUs) were applied to summarize state information from the lower encoder blocks. Although the MSC with 6 layers (base) achieves 27.68 BLEU with 73M parameters, the performance is lower than MPT’s 28.4 BLEU while requiring 1.2 times larger memory than MPT's 61.2M parameters.

\section{Discussion}
Our main goal is to maximize performance without increasing memory consumption, and we show that MPT can achieve comparable performance with SOTA methods while having a much lower memory consumption. 

Although much deeper models can achieve higher BLEU scores, more memory consumption is required.  Even if a 12-layer BT could achieve similar performance, MPT would still be better than such a deeper BT in terms of memory consumption as shown in Table \ref{tab:result_sota}. To compare MPT with deeper networks, a fully unconstrained 12-layer BT and with a 6+6-layer model simply concatenating two 6-layer BTs with tied weights (but without feeding the original input to the 7th layer nor allowing extra direct connections between layers 1-6 and 7-12 as we propose) were tested on EN-DE. The unconstrained 12-layer BT obtained comparable performance to our method, with 28.5 BLEU, but using twice larger memory consumption for the encoder. The concatenated 6+6 layer model with tied weights obtained 27.8 without an improvement. 

The best number of passes for MPT needs to be examined.  We did evaluate a 3-pass transformer, and found it performed comparably with 2-pass on EN-DE, with 28.4 BLEU. We applied 2-pass to all experiments in the paper. MPT can be applied to other tasks such as automatic speech recognition and the best number of passes may be task-dependent. 

\section{Conclusion}
We proposed a multi-pass transformer for efficient multi-stage fusion of self-attention features, %
and explored both soft and hard connection variants, using random architecture search for the latter.
Multi-pass transformer can achieve comparable performance with Large Transformer while using the same number of parameters as Base Transformer. Our proposed multi-pass architecture can also be used with other architectures such as LSTM and CNN. Future work includes testing the performance of our approach on new architectures and tasks such as language understanding. We will also explore other orthogonal research directions such as efficient multiple layer aggregation, dynamic routing, and dense connection. 
\bibliography{anthology,acl2020}
\bibliographystyle{acl_natbib}

\end{document}